\documentclass[acmtog]{acmart}
\acmSubmissionID{1325}

\usepackage{booktabs} % For formal tables

\newcommand{\revist}[1]{#1}

\usepackage[ruled]{algorithm2e} % For algorithms

\SetAlFnt{\small}
\SetAlCapFnt{\small}
\SetAlCapNameFnt{\small}
\SetAlCapHSkip{0pt}

\acmJournal{TOG}
\usepackage{booktabs}
\usepackage{tabularx}
\usepackage{multirow}
\usepackage{pifont}
\newcommand{\nameofmethod}{Mutual Forcing}
\newcommand{\figref}[1]{Fig.~\ref{#1}}
\newcommand{\tabref}[1]{Tab.~\ref{#1}}

\newcommand{\cmark}{\ding{51}}%
\newcommand{\xmark}{\ding{55}}%

\copyrightyear{2026}
\acmYear{2026}
\setcopyright{cc}
\setcctype{by-nc-nd}
\acmConference[SA Conference Papers '26]{SIGGRAPH Asia 2026 Conference Papers}{December 01--04, 2026}{Kuala Lumpur, Malaysia}
\acmBooktitle{SIGGRAPH Asia 2026 Conference Papers (SA Conference Papers '26), December 01--04, 2026, Kuala Lumpur, Malaysia}
\acmDOI{10.1145/3829340.3842196}
\acmISBN{979-8-4007-2842-6/2026/12}

\begin{document}

\title{Mutual Forcing: Dual-Mode Self-Evolution for Fast Autoregressive Audio-Video Character Generation}

\author{Yupeng Zhou}
\orcid{0009-0005-5113-2796}
\affiliation{%
  \institution{VCIP, School of Computer Science, Nankai University}
  \city{Tianjin}
  \country{China}
}
\email{ypzhousdu@gmail.com}

\author{Lianghua Huang}
\orcid{0000-0002-9686-9354}
\affiliation{%
  \institution{Wan Team, Alibaba Group}
  \city{Beijing}
  \country{China}
}
\email{xuangen.hlh@alibaba-inc.com}

\author{Zhi-Fan Wu}
\orcid{0009-0006-8417-165X}
\affiliation{%
  \institution{Wan Team, Alibaba Group}
  \city{Beijing}
  \country{China}
}
\email{wuzhifan.wzf@alibaba-inc.com}

\author{Jiabao Wang}
\orcid{0000-0003-4951-3088}
\affiliation{%
  \institution{VCIP, School of Computer Science, Nankai University}
  \city{Tianjin}
  \country{China}
}
\email{jbwang@mail.nankai.edu.cn}

\author{Yupeng Shi}
\orcid{0009-0007-1719-4293}
\affiliation{%
  \institution{Wan Team, Alibaba Group}
  \city{Beijing}
  \country{China}
}
\email{shiyupeng.syp@alibaba-inc.com}

\author{Biao Jiang}
\orcid{0009-0008-6051-1183}
\affiliation{%
  \institution{School of Intelligence Science and Technology, Peking University}
  \city{Beijing}
  \country{China}
}
\email{bjiang25@stu.pku.edu.cn}

\author{Daquan Zhou}
\orcid{0000-0002-4771-1796}
\affiliation{%
  \institution{School of Electrical and Computer Engineering, Peking University}
  \city{Shenzhen}
  \country{China}
}
% Replace with the exact email used in ACM Data Validation.
\email{zhoudaquan21@gmail.com}

\author{Yu Liu}
\orcid{0000-0001-8071-3745}
\affiliation{%
  \institution{Wan Team, Alibaba Group}
  \city{Beijing}
  \country{China}
}
\email{ly103369@alibaba-inc.com}

\author{Ming-Ming Cheng}
\orcid{0000-0001-5550-8758}
\affiliation{%
  \institution{VCIP, School of Computer Science, Nankai University}
  \city{Tianjin}
  \country{China}
}
\affiliation{%
  \institution{NKIARI}
  \city{Shenzhen}
  \country{China}
}
\email{cmm@nankai.edu.cn}

\author{Qibin Hou}
\orcid{0000-0002-8388-8708}
\authornote{Corresponding author.}
\affiliation{%
  \institution{VCIP, School of Computer Science, Nankai University}
  \city{Tianjin}
  \country{China}
}
\affiliation{%
  \institution{NKIARI}
  \city{Shenzhen}
  \country{China}
}
\email{houqb@nankai.edu.cn}

\begin{abstract}
In this work, we propose \textbf{\nameofmethod{}}, a framework for fast autoregressive audio-video generation with long-horizon audio-video synchronization.
Our approach addresses two key challenges: joint audio-video modeling and fast autoregressive generation.
To ease joint audio-video optimization, we adopt a two-stage training strategy: we first train uni-modal generators and then couple them into a unified audio-video model for joint training on paired data.
For streaming generation, we ask whether a native fast causal audio-video model can be trained directly, instead of following existing streaming distillation pipelines that typically train a bidirectional model first and then convert it into a causal generator through multiple distillation stages.
Our answer is Mutual Forcing, which builds directly on a native autoregressive model and integrates few-step and multi-step generation within a single weight-shared model, enabling self-distillation and improved training-inference consistency. 
The multi-step mode improves the few-step mode via self-distillation, while the few-step mode generates historical context during training to improve training-inference consistency; because the two modes share parameters, these two effects reinforce each other within a single model.
 Compared with prior approaches such as Self-Forcing, \nameofmethod{} removes the need for an additional bidirectional teacher model, supports more flexible training sequence lengths, reduces training overhead, and allows the model to improve directly from real paired data rather than a fixed teacher. Experiments show that \nameofmethod{} matches or surpasses strong baselines that require around 50 sampling steps while using only 4 to 8 steps, demonstrating substantial advantages in both efficiency and quality.
\end{abstract}

\begin{teaserfigure}
    \centering
    \includegraphics[width=0.98\textwidth]{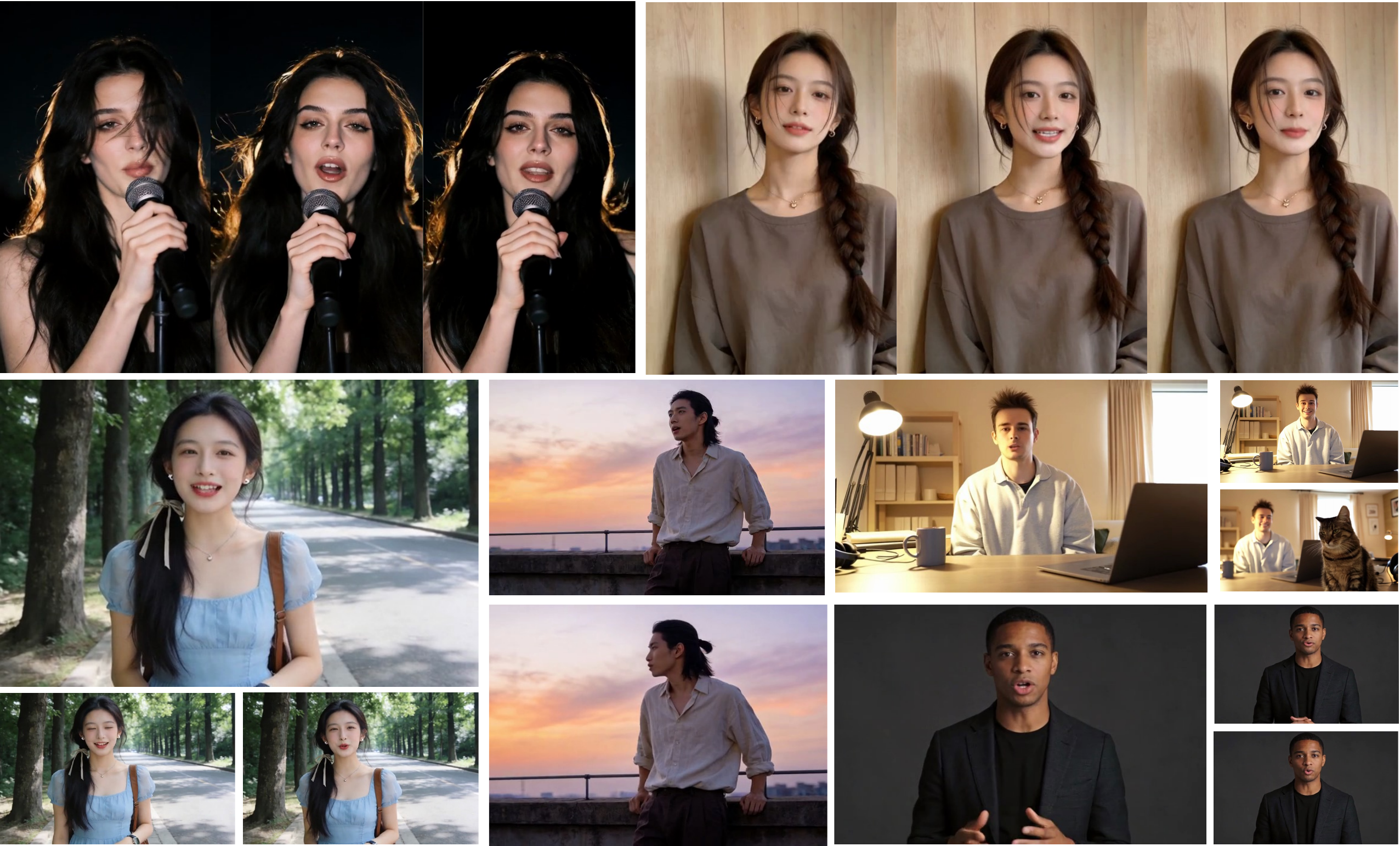}
\caption{Diverse audio-video character generation results produced by Mutual Forcing.}
    \label{fig:top_teaser}
\end{teaserfigure}




\maketitle

\begin{figure*}[th]
    \centering
    \includegraphics[width=1\textwidth]{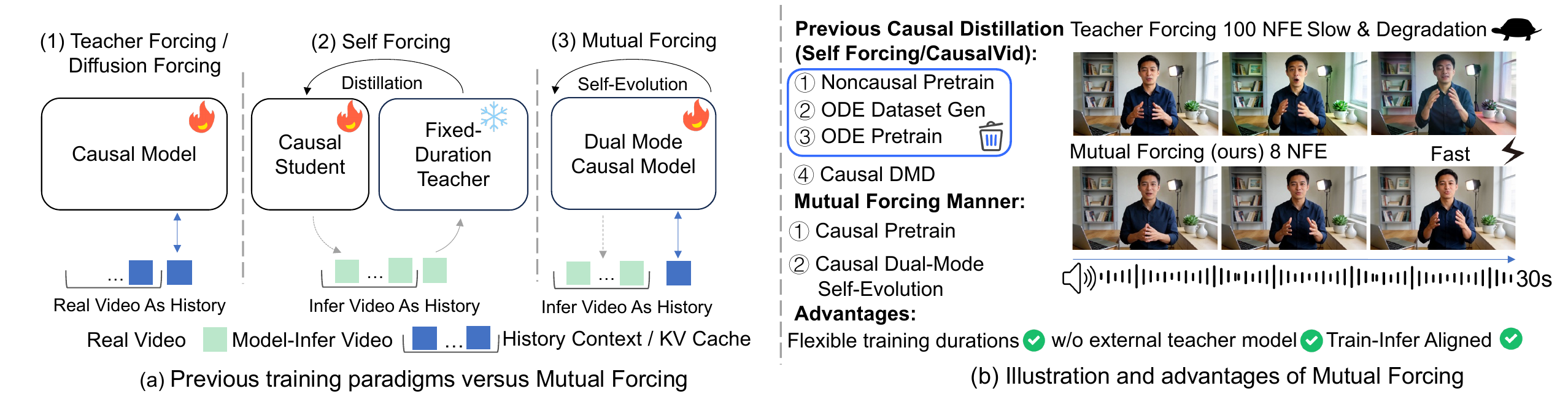}
% \caption{Illustration of Mutual Forcing.}
\caption{Illustration of \nameofmethod{}. 
(a) Comparison between prior training paradigms and \nameofmethod{}. 
Teacher forcing / diffusion forcing learns from real video history but suffers from training--inference mismatch. 
Self-Forcing replaces real history with model-inferred history, improving alignment but requiring an additional fixed-duration bidirectional teacher. 
\nameofmethod{}, by contrast, starts from a native causal model and employs a weight-shared dual-mode design to unify few-step and multi-step generation, enabling self-distillation, external-teacher-free training, flexible training sequence lengths, and training--inference consistency. 
(b) Qualitative results and key advantages of \nameofmethod{}. 
On long-duration streaming audio-video generation, \nameofmethod{} produces stable results with only 8 NFEs, compared with 100 NFEs for a teacher-forcing-trained baseline.}
    \label{fig:teaser}
\end{figure*}
\section{Introduction}

Existing research has focused mainly on conditional generation tasks that operate within a single modality, such as text-to-video \cite{wan2025wan,kong2024hunyuanvideo,hong2022cogvideo}, image-to-video~\cite{ren2024consisti2v,niu2024mofa,jin2024pyramidal}, and audio-to-video~\cite{gan2025omniavatar,gao2025wans2v,wang2025fantasytalking} generation.
The native joint generation of videos with synchronized audio has not yet been thoroughly explored.
Many previous works~\cite{ruan2023mm,wang2025av,haji2025av,ishii2025simple,liu2025javisdit} have made early attempts at joint audio-video generation.
However, they are typically limited to narrow-domain datasets or simple sounds.
Motivated by rapid progress in video generation and the impressive audio-video generation results from closed-source models, we explore the joint generation of audio-video on a larger scale in this work, in parallel with several recent concurrent works~\cite{low2025ovi,wang2025universe,huang2025jova,zhang2025uniavgen}.

Training a joint audio-video model from scratch poses a significant optimization challenge, as the model must jointly satisfy two tightly coupled objectives: (i) maintaining semantic fidelity to the text condition, and (ii) achieving precise audio-video synchronization.
These constraints are strongly intertwined, often making the early training signal unstable and leading to slow and suboptimal convergence. 
We first train an audio generator and a video generator separately to obtain well-formed single-modal models.
These models are subsequently integrated for joint training.
To facilitate seamless fusion, we keep the two branches architecturally identical and enable cross-modal interaction by coupling their self-attention, allowing audio and video tokens to attend to each other within the same attention computation.

Meanwhile, unlike concurrent works~\cite{huang2025jova,zhang2025uniavgen} that focus on fixed-length (e.g., 5-second) generation via 50-step bidirectional diffusion sampling, we explore a fast autoregressive paradigm for audio-video joint generation.
Formally, our model learns the conditional distribution $p_\theta(\mathbf{x}_t \mid \mathbf{x}_{<t})$ to generate the next frame $\mathbf{x}_t$ given previous frames $\mathbf{x}_{<t}$. 
This setting introduces two additional challenges: accelerating sampling and mitigating autoregressive degradation.
Further, we ask a more fundamental question: instead of following existing streaming distillation pipelines~\cite{huang2025self,yin2025slow} that typically begin with a bidirectional model and then convert it into a causal generator through multiple distillation stages, can we train a native fast causal audio-video model directly and then endow it with few-step generation ability?
We propose \nameofmethod{}, an external-teacher-free method that adopts a dual-mode design, enabling the model to support both few-step and multi-step generation. Furthermore, the interaction between the two modes facilitates self-evolution during training (Fig.~\ref{fig:teaser}(a)).
\nameofmethod{} consists of two optimization objectives:  a train-inference consistency objective and a self-distillation objective.
The train-inference consistency objective uses the few-step mode to infer the history frames and trains the multi-step mode on real frames with flow-matching loss, whereas the self-distillation objective trains the few-step mode using the multi-step mode as the teacher.
Because the dual modes share the same model parameters, these two intertwined objectives become mutually reinforcing. Thus, we term the approach \nameofmethod{}.

As shown in \figref{fig:teaser}(b), \nameofmethod{} offers clear advantages over prior training paradigms. Compared with teacher forcing or diffusion forcing, \nameofmethod{} achieves train--inference consistency and delivers a substantial speedup. Compared with self-forcing, \nameofmethod{} does not require an additional bidirectional teacher, supports more flexible supervision sequence lengths (instead of a fixed 5\,s), and learns directly from real data rather than being capped by the teacher model's performance.
Our contributions are summarized as follows:
\begin{itemize}
    \item We present a practical training recipe for large-scale audio-video generation and propose a two stage training scheme, which reduces the optimization difficulty, achieving strong performance among current research-purpose models.
    \item We explore autoregressive streaming generation and introduce a streaming text control mechanism, where a global caption specifies the overall scene while timestamped ASR tokens provide natural, fine-grained control over the evolving speech content.
    \item We introduce \nameofmethod{}, a self-distilled approach that integrates dual modes within a single model and drives self-evolution via two coupled optimization objectives, achieving step-reduced generation while mitigating streaming degradation, and supports training with flexible sequence lengths.
\end{itemize}

\section{Related Work}
\subsection{Conditional Video Generation}
Conditional video generation~\cite{wan2025wan,kong2024hunyuanvideo,gao2025wans2v,gan2025omniavatar,jin2024pyramidal,ma2024follow} synthesizes temporally coherent videos conditioned on diverse modalities, including text, audio, reference images, and structured motion (e.g., poses).
Early text-to-video diffusion models extended image diffusion to the temporal domain and established widely used conditioning mechanisms for video synthesis~\cite{ho2022video,ho2022imagen,he2022latent,wang2023modelscope}. 
Building upon latent diffusion~\cite{rombach2022high}, many works~\cite{guo2023animatediff,guo2024sparsectrl,xu2024magicanimate} adopted U-Net backbones with temporal modules and lightweight control components. 
Recently, to better scale model capacity and capture long-range spatiotemporal dependencies, video diffusion architectures have increasingly shifted from U-Nets to Transformers, a trend further popularized by Sora’s technical report~\cite{openai2024sora} and exemplified by recent large-scale models~\cite{yang2024cogvideox,wan2025wan,kong2024hunyuanvideo,jin2024pyramidal,zhou2024allegro}.

\subsection{Audio-Video Generation}

Audio-video generation can be divided into two-stage and joint methods.
Two-stage methods~\cite{gao2025wans2v,gan2025omniavatar,wang2025fantasytalking} first obtain audio and then generate an aligned video, typically by extending pretrained video models with audio-interaction modules.
Joint methods generate audio and video together in one model conditioned on prompts or other controls, enabling more flexible generation and cross-modal synchronization.
\revist{Existing audio-video data can be broadly divided into interaction-focused
and general web-video datasets. Seamless~\cite{agrawal2025seamless} contains
over 4,000 hours of purpose-collected face-to-face interactions, while
SpeakerVid-5M~\cite{zhang2025speakervid} curates over 8,700 hours of
speech-centric human videos from YouTube. In contrast,
InternVid~\cite{wang2023internvid} and Panda70M~\cite{chen2024panda}
provide much larger-scale general web-video collections with automatically
generated captions, offering broader visual diversity.}
Early works~\cite{ruan2023mm,haji2025av,ishii2025simple,wang2025av,liu2025javisdit} explored this setting but were often limited to narrow domains or simple sound events.
Motivated by recent progress in video generation and strong closed-source results, we study audio-video joint generation alongside recent and concurrent efforts~\cite{low2025ovi,wang2025universe,huang2025jova,zhang2025uniavgen}.
Compared with these works, our method focuses on longer-duration and faster streaming generation.

\subsection{Autoregressive Video Generation}
Autoregressive video generation synthesizes videos causally from past context, from early GAN-based methods~\cite{vondrick2017generating,ge2022long} to token-based autoregressive Transformers over discrete video codes~\cite{wang2024loong,wang2024emu3}. Since token-based models are costly due to the large number of tokens per frame, recent work has explored \emph{frame-based} autoregressive generation that directly predicts the next frame or chunk~\cite{yin2025slow,teng2025magi,bruce2024genie}. For online streaming generation under strict latency and no future-frame access, diffusion models face two key challenges: few-step sampling and exposure bias. Few-step acceleration typically relies on distillation, including DMD~\cite{yin2024improved}, consistency models~\cite{song2023improved}, and ShortCut~\cite{frans2024one}, with video extensions such as PhaseDMD~\cite{fan2025phased} and rCM~\cite{zheng2025large}; under streaming constraints, prior methods mainly adopt DMD-style distillation~\cite{yin2025slow,huang2025self,yang2025longlive,cui2025self}. Exposure bias stems from the mismatch between ground-truth training context and model-generated inference context, causing temporal drift, and has been addressed by noisy-context training~\cite{chen2024diffusion,chen2025skyreels,yin2025slow}, self-forcing~\cite{huang2025self,cui2025self}, and attention-sink windowed context~\cite{yang2025longlive}. However, Self-Forcing requires an extra bidirectional teacher and fixed-length supervision, while DMD-based methods remain expensive and are mostly validated on small video-only models. We therefore propose \nameofmethod{}, a streaming joint audio-video diffusion framework that unifies few-step and multi-step inference in one weight-shared model, enabling self-evolution training without an extra bidirectional teacher and scaling to large joint audio-video generation.

\section{Method}

\subsection{Problem Formulation}

\subsubsection{Audio-Video Joint Generation.}
The generative framework $f_\theta$ produces a video sequence and its temporally synchronized audio.
Let $c$ denote the control signal, which can be a text prompt, an image, or audio/video references.
We represent the target video and audio in latent token spaces as $\mathbf{v}_{1:T_v}$ and $\mathbf{a}_{1:T_a}$, where $T_v$ and $T_a$ are the token lengths of the two modalities.
Our goal is to learn a conditional mapping from \emph{pure noise} to clean audio-video latents.
Given clean latent tokens $(\mathbf{v}_{1:T_v}, \mathbf{a}_{1:T_a})$ and noise latents $(\mathbf{z}^v_{1:T_v}, \mathbf{z}^a_{1:T_a})$ sampled from a simple prior, the generator predicts
\begin{equation}
(\hat{\mathbf{v}}_{1:T_v},\,\hat{\mathbf{a}}_{1:T_a})
= f_\theta\!\left(\mathbf{z}^v_{1:T_v},\,\mathbf{z}^a_{1:T_a}\mid c\right),
\end{equation}
where the outputs are expected to be temporally aligned and cross-modally coherent.

We train the generator using flow matching.
Let $\mathbf{x}=(\mathbf{v}_{1:T_v},\mathbf{a}_{1:T_a})$ be clean audio-video latents and $\mathbf{z}\sim p(\mathbf{z})$ be pure noise.
We sample $t\sim\mathcal{U}(0,1)$ and define $\mathbf{x}_t=(1-t)\mathbf{z}+t\mathbf{x}$.
The model predicts a velocity field $v_\theta(\mathbf{x}_t,t,c)$ and is optimized with
\begin{equation}
\mathcal{L}_{\mathrm{FM}}
=\mathbb{E}\big[\|v_\theta(\mathbf{x}_t,t,c)-(\mathbf{x}-\mathbf{z})\|_2^2\big].
\end{equation}

\subsubsection{Streaming Generation.}
Most existing audio-video generators follow a non-streaming formulation: given $c$, they generate a fixed-length clip in one shot using non-causal full-context computation.
While effective for short clips, this formulation is less suitable for long-form generation because it requires a predefined horizon, scales memory and compute quadratically with clip length under full-context attention, and often degrades beyond the training window. In streaming generation, the model instead operates causally, generating audio and video incrementally while conditioning only on $c$ and previously generated outputs.
This enables low-latency, long-horizon synthesis.
We generate audio and video in chunks (one frame in our work).
Let $\mathcal{C}_k$ denote the $k$-th chunk, having a contiguous range of audio and video tokens.
At streaming step $k$, the model observes only past tokens and previously generated chunks, and predicts the next chunk as
\begin{equation}
p_\theta(\mathcal{C}_k \mid c, \mathcal{C}_{<k}), \qquad \text{with } \mathcal{C}_{<k}=\{\mathcal{C}_1,\ldots,\mathcal{C}_{k-1}\}.
\end{equation}
This causal factorization yields low-latency generation with linear memory and compute scaling, while requiring the model to maintain long-horizon consistency as $k$ grows.

\begin{figure}[!tp]
    \centering
    \includegraphics[width=\linewidth]{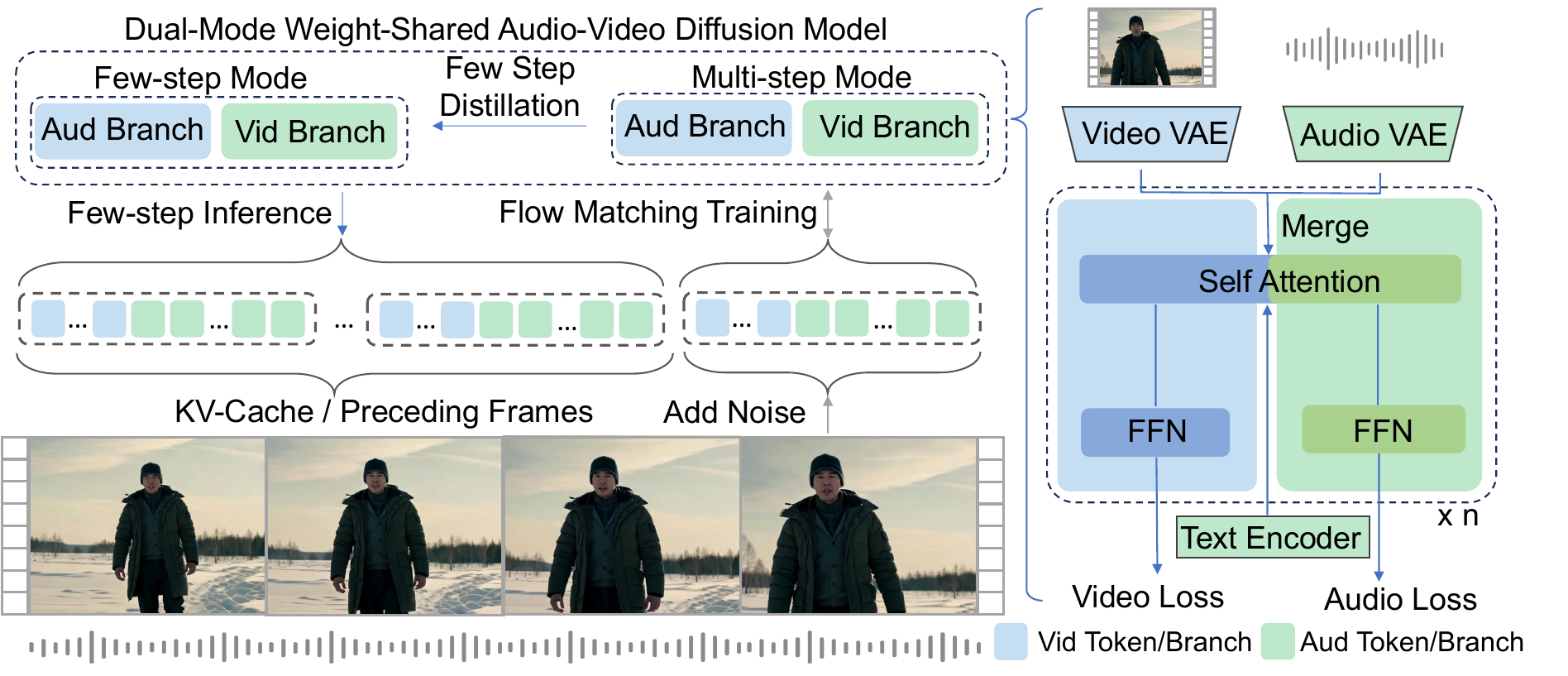}
\caption{Pipeline of \nameofmethod{} for streaming audio-video joint diffusion generation. Our model operates in two
weight-shared modes: a \textsc{Multi}-step mode and a \textsc{Few}-step mode, enabling self-evolution without a teacher. During \textsc{Multi}-step training, the \textsc{Few}-step mode generates the preceding tokens to form an inferred history; the \textsc{Multi}-step mode then predicts the next frame and is
supervised by the ground-truth target. Conversely, the \textsc{Few}-step mode is trained by distilling from the
\textsc{Multi}-step mode. The backbone uses modality-specific VAEs and modality-specific branches;
audio and video tokens are coupled and interact through shared self-attention.}
    \label{fig:pipeline}
\end{figure}

\subsection{Model Architecture and Training}

\subsubsection{Dual-branch Model.}
As shown on the right of \figref{fig:pipeline}, we adopt a unified
audio-video backbone with two structurally identical modality-specific
Transformer branches for audio and video. Each branch consists of
self-attention and feed-forward blocks. \revist{Their self-attention
computations are coupled, allowing audio, video, and text tokens to interact
directly within the same attention computation.}
\revist{We initialize the two branches through separate unimodal pretraining,
followed by teacher-forcing joint audio-video training and \nameofmethod{}
fine-tuning.}

\subsubsection{3D RoPE Embedding for Streaming Representations.}
To distinguish multimodal positional information, we introduce a 3D RoPE~\cite{su2021roformer} encoding that factorizes position into temporal, height, and width coordinates. We apply it to video, audio, and text tokens. For audio and text, the height and width coordinates are set to 0. All positions are computed from the actual timestamps of the corresponding audio, video, and text, ensuring temporal alignment across modalities.

\subsubsection{Two-stage Training Strategy.}
We adopt a two-stage training strategy to ease optimization under a coupled
audio-video generation objective.
\revist{In the first stage, we separately pretrain the audio and video branches
with modality-specific objectives to build strong unimodal priors.
In the second stage, we first perform teacher-forcing joint audio-video training
on paired data, followed by Mutual Forcing fine-tuning for efficient streaming
generation.}
\subsection{\nameofmethod{} for Fast Audio-Video Generation}
\label{sec:mutual_forcing}

\subsubsection{Dual-Mode Weight-Shared Model.}
We build \nameofmethod{} on flow matching and cast streaming prediction as a conditional diffusion ODE.
At each streaming step, the goal is to generate the next clean target conditioned on the previously observed or generated context $c$.
We introduce a diffusion-time variable $t\in[0,1]$ and a trajectory $\{x_t\}$ that maps Gaussian noise at $t=0$ to a clean sample at $t=1$.
Following prior work, we parameterize the trajectory with a context-conditioned velocity field $v_{\theta}(\cdot\mid c)$:
\begin{equation}
\label{eq:pf-ode-cond}
\frac{\mathrm{d}x_t}{\mathrm{d}t}=v_{\theta}(x_t,t \mid c), \qquad t\in[0,1],
\end{equation}
where $x_t\in\mathbb{R}^d$ is the state at diffusion time $t$.
The state transition is
\begin{equation}
\label{eq:ode-integrate-cond}
x_{t_2}=x_{t_1}+\int_{t_1}^{t_2} v_{\theta}(x_t,t \mid c)\,\mathrm{d}t,
\end{equation}
for any $0\le t_1<t_2\le 1$.

To support different inference budgets without multiple networks, we use a
dual-mode weight-shared model $M_{\theta}$: \textsc{Multi} uses small ODE
steps, while \textsc{Few} predicts large diffusion-time jumps.
\revist{The two modes share the same backbone and differ only in
mode-specific time conditioning. Specifically, we introduce an additional
zero-initialized time-embedding module: \textsc{Multi} uses the original
time embedding, while \textsc{Few} uses the sum of the original and
additional embeddings. This preserves the pretrained \textsc{Multi}-mode
behavior at initialization while enabling mode-specific adaptation.}

In \textsc{Multi} mode, the model receives $(x_t,t,c)$ and outputs the instantaneous velocity:
\begin{equation}
\label{eq:many-mode-cond}
M_{\theta}(x_t,t,c;\textsc{Multi}) = v_{\theta}(x_t,t \mid c).
\end{equation}

In \textsc{Few} mode, the model additionally receives the endpoint $t_2$ and predicts the interval-averaged velocity over $[t_1,t_2]$:
\begin{equation}
\label{eq:few-mode-cond}
M_{\theta}(x_{t_1},t_1,t_2,c;\textsc{Few})
\approx \frac{1}{t_2-t_1}\int_{t_1}^{t_2} v_{\theta}(x_t,t \mid c)\,\mathrm{d}t.
\end{equation}
The corresponding large-step update is
\begin{equation}
x_{t_2} \approx x_{t_1} + (t_2-t_1)\,
M_{\theta}(x_{t_1},t_1,t_2,c;\textsc{Few}).
\end{equation}
For self-distillation, the \textsc{Few}-mode target over $[t_1,t_2]$ is obtained by integrating stop-gradient \textsc{Multi}-mode predictions under the same context.

\subsubsection{Multi-step Mode Training Objective.}  
In \textsc{Multi} mode, we train $M_{\theta}$ to predict the instantaneous
velocity field under a self-evolving streaming context. Specifically, for
streaming step $k$, we construct the \revist{generated history $h_k$} by running
the same model in \textsc{Few} mode:
\begin{equation}
\label{eq:context-infer}
\begin{aligned}
\revist{\hat{C}_{k}}
&= \textsc{FewSample}\!\left(M_{\theta},\revist{h_{k-1}}\right),\\
\revist{h_{k}}
&= \operatorname{Update}\!\left(\revist{h_{k-1}},\hat{C}_{k}\right),
\end{aligned}
\end{equation}
where $\textsc{FewSample}(\cdot)$ denotes few-step ODE sampling with large time jumps, and
$\operatorname{Update}(\cdot)$ \revist{updates the generated
history} by incorporating the newly generated chunk.
\revist{The 4-NFE and 8-NFE models use 4-NFE and 8-NFE rollouts, respectively, to autoregressively construct the preceding history up to the sampled target step. The generated history and its cached representations are detached from the computation graph and used only as conditioning context; gradients are propagated only through the prediction of the current target chunk.}

As shown in \figref{fig:pipeline}, given this \revist{model-generated history}, we minimize the standard flow-matching
regression loss
\begin{equation}
\mathcal{L}_{\textsc{Multi}}(\theta)
=
\mathbb{E}_{k,t}\Big[
\big\|
M_{\theta}(x_t,t,\revist{h_k};\textsc{Multi}) - u_t
\big\|_2^2
\Big]
\end{equation}
where $k$ indexes the streaming step sampled from training sequences, $t\sim\mathcal{U}[0,1]$,
$x_t$ is the intermediate state at diffusion time $t$ constructed from the current training target
(e.g., the next frame/segment), and $u_t$ is the corresponding flow-matching target velocity.
Minimizing $\mathcal{L}_{\textsc{Multi}}$ encourages $M_{\theta}$ to match
the target velocity distribution under its own \revist{generated history}.

\subsubsection{Few-step training objective.}
As described above, we use \textsc{Few}-mode sampling both to construct self-evolving streaming
context for \textsc{Multi}-mode training and to enable fast streaming inference.
Instead of using two separate networks as in DMD-style distillation, we train a single dual-mode,
weight-shared model.
The \textsc{Few} mode is optimized by self-distillation: it learns to predict the interval displacement
$x_{t_2}-x_{t_1}$ over $[t_1,t_2]$, using targets produced by the same model in \textsc{Multi} mode
with gradients stopped.

In practice, two common objectives exhibit complementary trade-offs.
\emph{ShortCut}-style~\cite{frans2024one} objectives are usually stable and easy to optimize, but their performance
degrades when the sampling budget becomes extremely small (e.g., 4 steps).
DMD-style~\cite{yin2024improved} objectives often provide better few-step performance, but can be unstable, especially
for large-scale audio-video generation.
We therefore use a hybrid objective that combines the stability of ShortCut with the effectiveness of DMD:
\begin{equation}
\label{eq:few-hybrid}
\mathcal{L}_{\textsc{Few}}(\theta)
=
\lambda\,\mathcal{L}_{\textsc{Few}}^{\textsc{DMD}}(\theta)
+(1-\lambda)\,\mathcal{L}_{\textsc{Few}}^{\textsc{SC}}(\theta),
\end{equation}
\revist{where we set $\lambda=1/3$ in all main experiments; a sensitivity
analysis is provided in the supplementary material.}
Here, $\mathcal{L}_{\textsc{Few}}^{\textsc{DMD}}$ matches the \textsc{Few}-mode prediction to the
interval displacement induced by the stop-gradient \textsc{Multi}-mode trajectory, and
$\mathcal{L}_{\textsc{Few}}^{\textsc{SC}}$ enforces step consistency for large-jump prediction.
\revist{The \textsc{Multi}-mode trajectory is treated as a detached target,
while gradients are propagated through the \textsc{Few}-mode prediction.}
\revist{Unlike conventional DMD,} our method uses the internal \textsc{Multi}-step mode
$M_{\theta}(x_t,t,c;\textsc{Multi})$ as the teacher, removing the need for an external teacher network. Detailed definitions of $\mathcal{L}_{\textsc{Few}}^{\textsc{DMD}}$ and $\mathcal{L}_{\textsc{Few}}^{\textsc{SC}}$ are provided in the supplementary material.

\subsubsection{Dual-Mode Self-Evolution.}
Streaming audio-video generation typically requires reducing sampling steps to meet strict latency constraints.
Accordingly, fast streaming diffusion models (e.g., Self-Forcing~\cite{huang2025self} and
CausVid~\cite{yin2025slow}) are often obtained by distilling a multi-step teacher into a few-step student.
However, this paradigm has two limitations. First, the student is bounded by the teacher’s capability and the
supervision is often effective over relatively short horizons (e.g., a few seconds in the Self-Forcing
setting).
% Moreover, Self-Forcing requires an additional bidirectional teacher; and (ii) 
Second, keeping separate
teacher/student models incurs a huge training overhead for large audio-video generators.

Mutual Forcing is optimized with two coupled objectives,
$\mathcal{L}_{\textsc{Multi}}$ and $\mathcal{L}_{\textsc{Few}}$.
Here, $\mathcal{L}_{\textsc{Multi}}$ is optimized with supervision from
paired training data, continually improving the shared parameters used by
both modes, while $\mathcal{L}_{\textsc{Few}}$ distills from the
\textsc{Multi}-step mode with stop-gradient.
\revist{In practice, the two objectives are back-propagated separately,
and the gradient from $\mathcal{L}_{\textsc{Few}}$ is rescaled to match
the gradient norm of $\mathcal{L}_{\textsc{Multi}}$ before the two
gradients are combined for parameter updating.}

\begin{table*}[!thp]
\centering
\small
\setlength{\abovecaptionskip}{2pt}
\setlength{\tabcolsep}{8pt}
\renewcommand{\arraystretch}{1.0}

\caption{Quantitative comparison with \textbf{audio-driven} and \textbf{audio-video joint} generation baselines, including Fantasy-Talking~\cite{wang2025fantasytalking}, Omni-Avatar~\cite{gan2025omniavatar}, Wan-S2V~\cite{gao2025wans2v}, Universe-1~\cite{wang2025universe}, and Ovi~\cite{low2025ovi}. 
\textbf{NFE} denotes the number of denoising-network forward passes during sampling; with CFG, each denoising step takes two forward passes. 
\nameofmethod{} achieves better performance with fewer NFEs.}
\label{tab:quantitative_table}

\begin{tabular}{l c c c c c c  c c c c c c c}
\toprule
\multirow{2}{*}{\textbf{Method}} & \multirow{2}{*}{\textbf{NFE}} & \multirow{2}{*}{\textbf{AR}} &
\multicolumn{1}{c}{\textbf{Audio-Video}}  &
\multicolumn{7}{c}{\textbf{Audio}} &
\multicolumn{3}{c}{\textbf{Video}} \\
\cmidrule(lr){4-4}\cmidrule(lr){5-11}\cmidrule(lr){12-14}
& & &  LSE-C$\uparrow$ & WER$\downarrow$ &
FD$\downarrow$ & KL$\downarrow$  & CE$\uparrow$ & CU$\uparrow$ & PC$\downarrow$ & PQ$\uparrow$ &
MS$\uparrow$ & AS$\uparrow$ & ID$\uparrow$ \\
\midrule

\multicolumn{14}{l}{\textit{Audio-driven Generation}}\\
Fantasy-Talking & 60 & \xmark & 2.48 & $-$  & $-$  &$-$  & $-$  & $-$  &$-$  &$-$ & 0.23 & 0.38 & \underline{0.87} \\
Omni-Avatar & 100 & \xmark & 6.07 & -- & -- & -- & --  & -- & -- & -- & 0.45 &  0.42& 0.81 \\
Wan-S2V       & 100  & \xmark  & 5.20  & -- & --  & -- & -- & -- & -- & -- & 0.54 & 0.40 & 0.85\\
\midrule

\multicolumn{14}{l}{\textit{ Audio-Video Joint Generation}}\\
Universe-1 &100 &\xmark & 6.01  & 0.26 & 0.48 & 0.45  & 3.61 & 3.64 & 1.80 & 4.06 & 0.38 & 0.41 & 0.85\\
Ovi & 100 & \xmark & \underline{6.19} & \underline{0.17} & 0.77 & 0.27 & 5.21 & 5.69 & 1.67 & 5.61 & \underline{0.55} & 0.42 & \textbf{0.88} \\
\textbf{\nameofmethod{}} & \textbf{4} & \cmark & 5.26 & {0.23} & \textbf{0.28} & \textbf{0.16}  & \underline{5.66} &  \underline{6.29} &  \underline{1.64} & \underline{6.44} & \textbf{0.59} & \underline{0.45} & 0.84  \\
\textbf{\nameofmethod{}} & \textbf{8} & \cmark & \textbf{6.35} & \textbf{0.11} & \underline{0.38} & \underline{0.21} & \textbf{5.77} & \textbf{6.51} & \textbf{1.61} & \textbf{6.83} & 0.37 & \textbf{0.47} & \textbf{0.88}  \\

\bottomrule
\end{tabular}
\end{table*}

\begin{figure*}[!th]
    \centering
    \includegraphics[width=\linewidth]{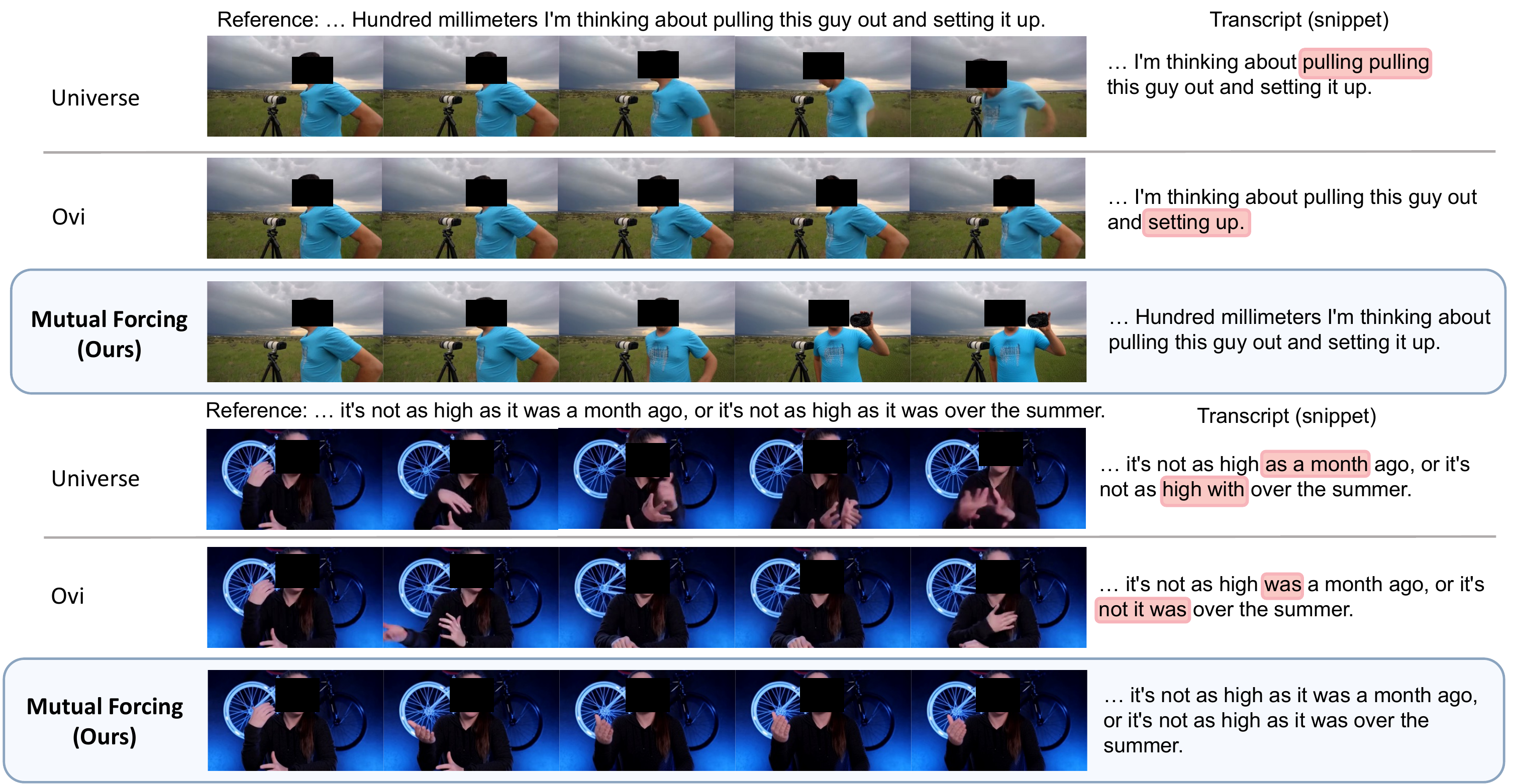}
    \caption{Qualitative comparison with Audio-Video Joint Generation Model Universe-1~\cite{wang2025universe} and Ovi~\cite{low2025ovi}. We visualize generated frame sequences and the corresponding speech transcripts. \nameofmethod{} (Ours) produces more accurate spoken content while maintaining coherent and temporally consistent visual progression, despite using substantially fewer sampling steps.}
    \label{fig:visual_comparison}
\end{figure*}

\begin{figure*}[!th]
    \centering
    \includegraphics[width=0.83\linewidth]{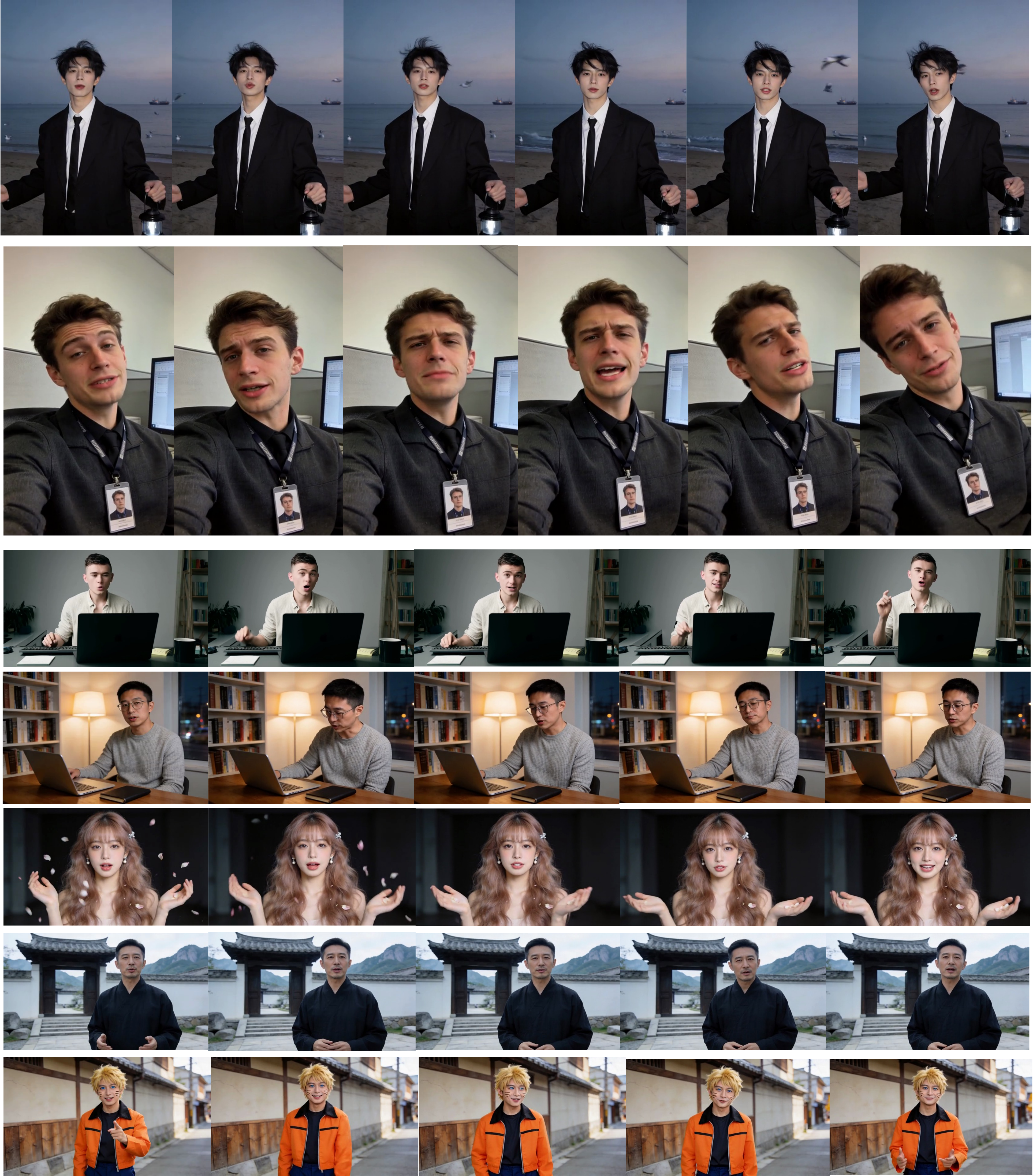}
\caption{Additional results generated by Mutual Forcing. 
Our method produces synchronized audio-video character animations with coherent motion, stable identity, and natural visual dynamics across diverse scenes.}
    \label{fig:visual_self}
\end{figure*}

\section{Experiments}

\subsection{Implementation Details}

Our model comprises two modality-specific Transformer branches for audio and
video, each containing 7B parameters, resulting in 14B parameters in total.
Each branch consists of 28 Transformer blocks with a hidden dimension of 3,584
and an MLP hidden dimension of 18,944. The self-attention module uses 28
attention heads with a head dimension of 128. We adopt Qwen2.5-VL 8B as the
text encoder.

For unimodal pretraining, the audio branch is trained on
Emilia~\cite{he2024emilia}, while the video branch uses diverse video data from
Panda70M~\cite{chen2024panda} and InternVid~\cite{wang2023internvid}.
For joint audio-video training, we focus on character-centric clips from
Seamless~\cite{agrawal2025seamless}, SpeakerVid-5M~\cite{zhang2025speakervid},
InternVid, and Panda70M. Our data filtering pipeline is mainly inspired by the
curation procedure of SpeakerVid-5M~\cite{zhang2025speakervid}; for Panda70M
and InternVid, we additionally apply person/face detection and SyncNet-based
audio-video consistency filtering. Training clips range from 1.5 to 10 seconds,
and the autoregressive chunk size is one frame. At a resolution of
$192\times384$, each video frame contains approximately 0.3K latent tokens.

During inference, generation is performed in a frame-wise autoregressive
manner. We use the Euler solver with a uniformly spaced flow-time grid and a
time shift of 3. For evaluation, we use 50 prompts covering diverse scenarios,
including multiple speakers, fast motion, camera motion, and non-speaking
sounds. More details are provided in the supplementary material.

\subsection{Quantitative Comparison}
Following Universe-1~\cite{wang2025universe} and JoVA~\cite{huang2025jova},
we evaluate audio-video alignment with SyncNet confidence (LSE-C)~\cite{chung2016out};
video quality with VBench~\cite{huang2024vbench} Motion Score (MS), Aesthetic
Score (AS), and identity consistency (ID); and audio quality with CLAP-based
KL/FD~\cite{laion2022clap}, AudioBox-Aesthetics (PQ/PC/CE/CU)~\cite{tjandra2025aes},
and SenseVoice-based WER~\cite{an2024funaudiollm}.
As shown in \tabref{tab:quantitative_table}, \nameofmethod{} achieves
better performance on most metrics with only 4 or 8 NFEs,
compared with 100 NFEs for the main baselines. NFE denotes the number of
denoising network evaluations. CFG requires two evaluations per sampling
step, while our few-step model does not use CFG at inference.

\subsection{Qualitative Comparison}
As shown in \figref{fig:visual_comparison}, we provide a qualitative comparison against joint audio-video generation models, Universe-1~\cite{wang2025universe} and Ovi~\cite{low2025ovi}.
Consistent with the quantitative results in \tabref{tab:quantitative_table}, \nameofmethod{} produces spoken content that better matches the intended lexical and phonetic structure, while preserving coherent visual dynamics and temporal continuity across frames. 
We further present additional generation results in \figref{fig:visual_self}, covering diverse characters, appearances, and scene configurations.
These examples demonstrate that \nameofmethod{} consistently generates synchronized audio-video content with stable identities, coherent motion, and natural visual dynamics across diverse scenarios.
Importantly, \nameofmethod{} achieves these gains with fewer sampling steps, further demonstrating the effectiveness of our approach.

\subsection{Ablation Study}

\subsubsection{\revist{Effect of Mutual Forcing Components.}}
\revist{We compare three variants as shown in Table~\ref{tab:history_ablation}.
Without Mutual Forcing, the model is trained with ground-truth histories.
Mutual Forcing with Multi-step histories introduces inferred-history training
using expensive Multi-step rollouts, leading to clear performance improvements.
The full Mutual Forcing model instead uses efficient Few-step histories and
achieves comparable performance with $12.5$--$25\times$ lower rollout cost.
These results show that Mutual Forcing substantially benefits from
inferred-history training, while Few-step rollouts provide an efficient
alternative to costly Multi-step history generation.}

\subsubsection{Analysis of the Weight-Shared Design.}
To understand the role of the weight-shared design in \nameofmethod{}, we analyze the attention behaviors of the few-step and multi-step modes to better understand why this design is effective. Since the two modes share the same model parameters, successful training requires them to develop compatible internal dynamics despite operating under different generation schedules. Fig.~\ref{fig:ablation_attention}(a) shows that the attention maps of the two modes are highly consistent across all layers, with similarity above 97\%. This strong agreement indicates that the self-evolution strategy effectively aligns the two modes at the representation level, allowing the few-step mode to inherit the robust generation of the multi-step mode without introducing a separate teacher model.

\subsubsection{Analysis of Temporal Attention Distribution.}
\begin{figure}[!th]
    \centering
    \includegraphics[width=\linewidth]{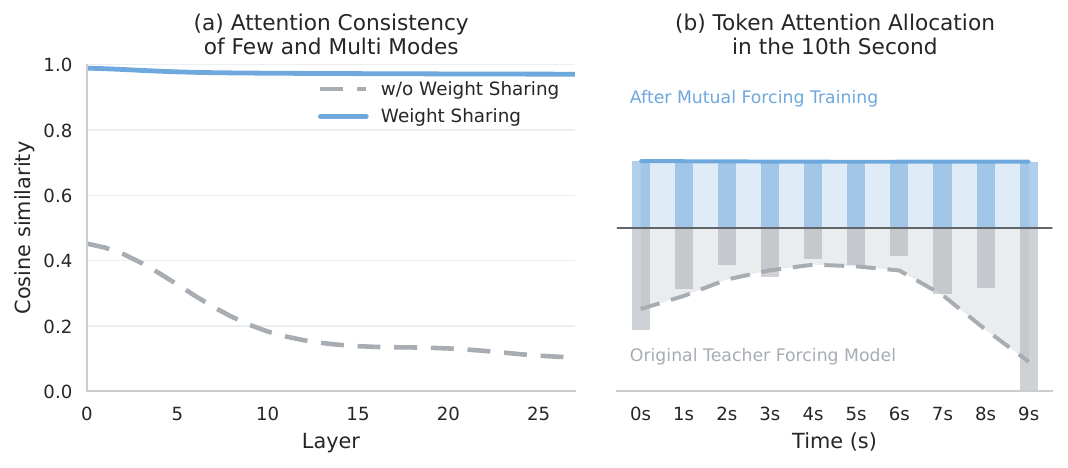}
\caption{Attention analysis of \nameofmethod{}. 
(a) Attention consistency between the few and multi modes across layers. 
The consistently high similarity (over 97\%) indicates that using few-mode inferred histories to train the multi mode successfully aligns the attention behaviors of the two modes, explaining the effectiveness of \nameofmethod{} in mitigating degradation. 
(b) Token attention allocation in the 10th second. 
Compared with the original teacher forcing model, \nameofmethod{} leads to a more balanced temporal attention distribution, reducing over-reliance on a few critical frames and thus alleviating progressive degradation.}
    \label{fig:ablation_attention}
\end{figure}

In Fig.~\ref{fig:ablation_attention}(b), we compare token attention allocation when using a token in the 10th second as the query. The original teacher-forcing model tends to place disproportionate attention on a small number of past frames, which may amplify errors when these frames become unreliable during inference. In contrast, \nameofmethod{} produces a more balanced temporal attention distribution, suggesting that the model learns to aggregate information from a broader historical context. This more stable attention behavior helps mitigate error accumulation and progressive degradation in long-duration generation.

\subsubsection{Human Evaluation.}
\begin{figure}[!t]
    \centering
    \includegraphics[width=\linewidth]{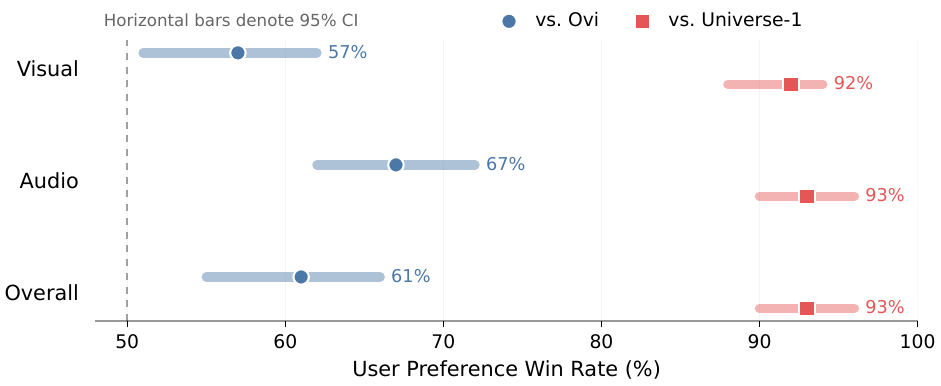}
    \caption{\revist{\textbf{Human evaluation results} against Ovi and Universe-1 on three criteria: visual preference, audio alignment, and overall quality. Points indicate the win rates of our method, and error bars denote the corresponding 95\% confidence intervals.}}
    \label{fig:user_study}
\end{figure}

To evaluate audio-video synchronization and perceived quality, we added a human preference study against Ovi and Universe-1.
\revist{The evaluation is randomized and blinded: participants compare outputs from two models without knowing their identities and select the better one or a tie. For each prompt, we evaluate three dimensions: visual preference, audio alignment, and overall quality. Each dimension of each prompt is evaluated by approximately 7 raters.}
As shown in Fig.~\ref{fig:user_study}, \nameofmethod{} obtains consistently higher win rates across visual preference, audio alignment, and overall quality.
In particular, the margin over Universe-1 is substantial across all three criteria, and \nameofmethod{} also maintains clear advantages over Ovi.
Overall, the human evaluation confirms that the gains of \nameofmethod{} are not only reflected in automatic metrics, but are also evident in human judgment.

\begin{table}[!t]
\centering
\small
\setlength{\abovecaptionskip}{2pt}
\setlength{\tabcolsep}{1.5pt}
\caption{\revist{
Ablation of Mutual Forcing and inferred-history training.
We compare the original Multi-step training without Mutual Forcing,
a variant using costly Multi-step inferred histories,
and the full Mutual Forcing model with Few-step histories.
Introducing inferred-history training clearly improves overall performance,
while Few-step rollouts achieve comparable results to costly Multi-step history generation at substantially lower rollout cost.
}}
\label{tab:history_ablation}
{
\begin{tabular}{@{}lccccc@{}}
\toprule
\textbf{Variant}
& \textbf{Inferred Hist.}
& \textbf{CE$\uparrow$}
& \textbf{CU$\uparrow$}
& \textbf{AS$\uparrow$}
& \textbf{ID$\uparrow$} \\
\midrule

w/o Mutual Forcing
& --
& 5.36
& 6.13
& 0.45
& 0.79 \\

Mutual Forcing w/ Multi-step History
& \checkmark
& \textbf{5.74}
& 6.40
& 0.45
& \textbf{0.85} \\

Mutual Forcing
& \checkmark
& 5.70
& \textbf{6.42}
& \textbf{0.46}
& \textbf{0.85} \\
\bottomrule
\end{tabular}}
\end{table}

\begin{table}[!t]
\centering
\small
\setlength{\abovecaptionskip}{2pt}
\setlength{\tabcolsep}{6pt}
\renewcommand{\arraystretch}{1.05}
\caption{\textbf{Long-horizon quality over time.} We report audio (CU/CE) and video (AS/ID) metrics over three temporal windows. \nameofmethod{} maintains more consistent quality over time than baselines.}
\begin{tabular}{llccccc}
\toprule
Method & Teacher & Window & CE$\uparrow$ & CU$\uparrow$ & AS$\uparrow$ & ID$\uparrow$ \\
\midrule
\multirow{3}{*}{DMD} 
& \multirow{3}{*}{Causal}
& 0--5s   & 5.25 & 6.46 & 0.39 & 0.79 \\
& & 5--15s  & 4.90 & 6.28 & 0.25 & 0.64 \\
& & 15--25s & 3.92 & 5.67 & 0.22 & 0.47 \\
\midrule
\multirow{3}{*}{ShortCut}
& \multirow{3}{*}{Causal}
& 0--5s   & 5.38 & 5.73 & 0.46 & 0.84 \\
& & 5--15s  & 5.17 & 5.61 & 0.39 & 0.78 \\
& & 15--25s & 4.94 & 5.47 & 0.32 & 0.64 \\
\midrule
\multirow{3}{*}{Self-Forcing}
& \multirow{3}{*}{Causal}
& 0--5s   & 4.37 & 5.11 & 0.43 & 0.80 \\
& & 5--15s  & 3.43 & 3.66 & 0.40 & 0.68 \\
& & 15--25s & 2.78 & 3.03 & 0.38 & 0.62 \\
\midrule
\multirow{3}{*}{\nameofmethod{}}
& \multirow{3}{*}{-}
& 0--5s   & 5.70 & 6.42 & 0.46 & 0.85 \\
& & 5--15s  & 5.75 & 6.54 & 0.46 & 0.84 \\
& & 15--25s & 5.41 & 6.38 & 0.46 & 0.85 \\
\bottomrule
\end{tabular}
\label{tab:long_video}
\end{table}

\subsubsection{Long-Video Inference Comparison.}
Although our dual-mode self-evolution is not trained on very long clips, we evaluate on 25-second videos and report metrics over 0--5\,s, 5--15\,s, and 15--25\,s; see \tabref{tab:long_video}.
\revist{All methods are trained on videos no longer than 10 seconds.}
Since no public long audio-video streaming baseline exists in our setting, we implement three baselines using conventional distillation with a streaming teacher: (i) DMD with teacher forcing, (ii) ShortCut with teacher forcing, and (iii) DMD with self-forcing. For (iii), the teacher context is built from ground-truth video, and teacher denoising of student-generated samples provides score distillation targets, which reduces long-horizon degradation and helps stabilize training with a causal teacher.
As shown in \tabref{tab:long_video}, \nameofmethod{} remains consistent across all temporal windows and generalizes well to 25-second inference despite not training on such long sequences. We attribute this to dual-mode self-evolution, which exposes the model to partially self-generated contexts during training and thus mitigates exposure bias and prediction drift. In contrast, all baselines degrade progressively over time in both video and audio quality. So, \nameofmethod{} is more robust for long-horizon streaming generation.

\subsubsection{Inference Time.}
We compare generation throughput with prior joint audio-video baselines, Ovi and Universe-1.
\revist{We measure throughput with a batch size of 1 using BF16 precision on NVIDIA GPUs with 80\,GB memory. For each comparison, we fix the resolution, sequence length, and video duration, and apply the same timing protocol to the baselines. The reported inference time excludes warm-up, data loading, model loading, video loading, and result saving.}
Since both baselines rely on bidirectional diffusion and do not support streaming autoregressive generation, we report the practical throughput of each method under its corresponding inference configuration, where the primary difference lies in the number of GPUs used at inference time. As shown in \tabref{tab:speed}, Universe-1 and Ovi achieve 0.6 FPS and 1.3 FPS, respectively, while requiring 4 and 8 GPUs. In contrast, \nameofmethod{} runs on a single GPU and reaches 30 FPS at $192\times336$, 12 FPS at $480\times768$, and 3.5 FPS at $704\times1280$. These results show that \nameofmethod{} achieves substantially higher practical throughput and supports real-time generation at low resolution as well as fast-streaming generation at higher resolutions.

\begin{table}[!t]
\centering
\small
\setlength{\abovecaptionskip}{2pt}
\setlength{\tabcolsep}{8.5pt}
\caption{Generation speed comparison under the same GPU type. 
All methods are evaluated using the same GPU model, while the number of GPUs follows each method's standard inference setup.}

\renewcommand{\arraystretch}{1.05}
\begin{tabular}{l c c  c c}
\toprule
\textbf{Method} & \textbf{Res.} & \textbf{Device}  &  \textbf{FPS} & \textbf{Speed}  \\
\midrule
Universe-1 &480$\times$768   &  4 GPUs      &  0.6   & Slow \\
Ovi  & 704$\times$1280         &  8 GPUs    &   1.3         & Slow \\
\midrule
\nameofmethod{} & 192$\times$336  & \textbf{1 GPU} & \textbf{30} &  \textbf{Real-time} \\ % 4
\nameofmethod{} & 480$\times$768  & \textbf{1 GPU} & \textbf{12} & \textbf{Fast} \\ % 10
\nameofmethod{} & 704$\times$1280  & \textbf{1 GPU} & \textbf{3.5} & \textbf{Fast} \\ % 34
\bottomrule
\end{tabular}

\label{tab:speed}
\end{table}

\begin{figure}[t]
    \centering
    \includegraphics[width=\linewidth]{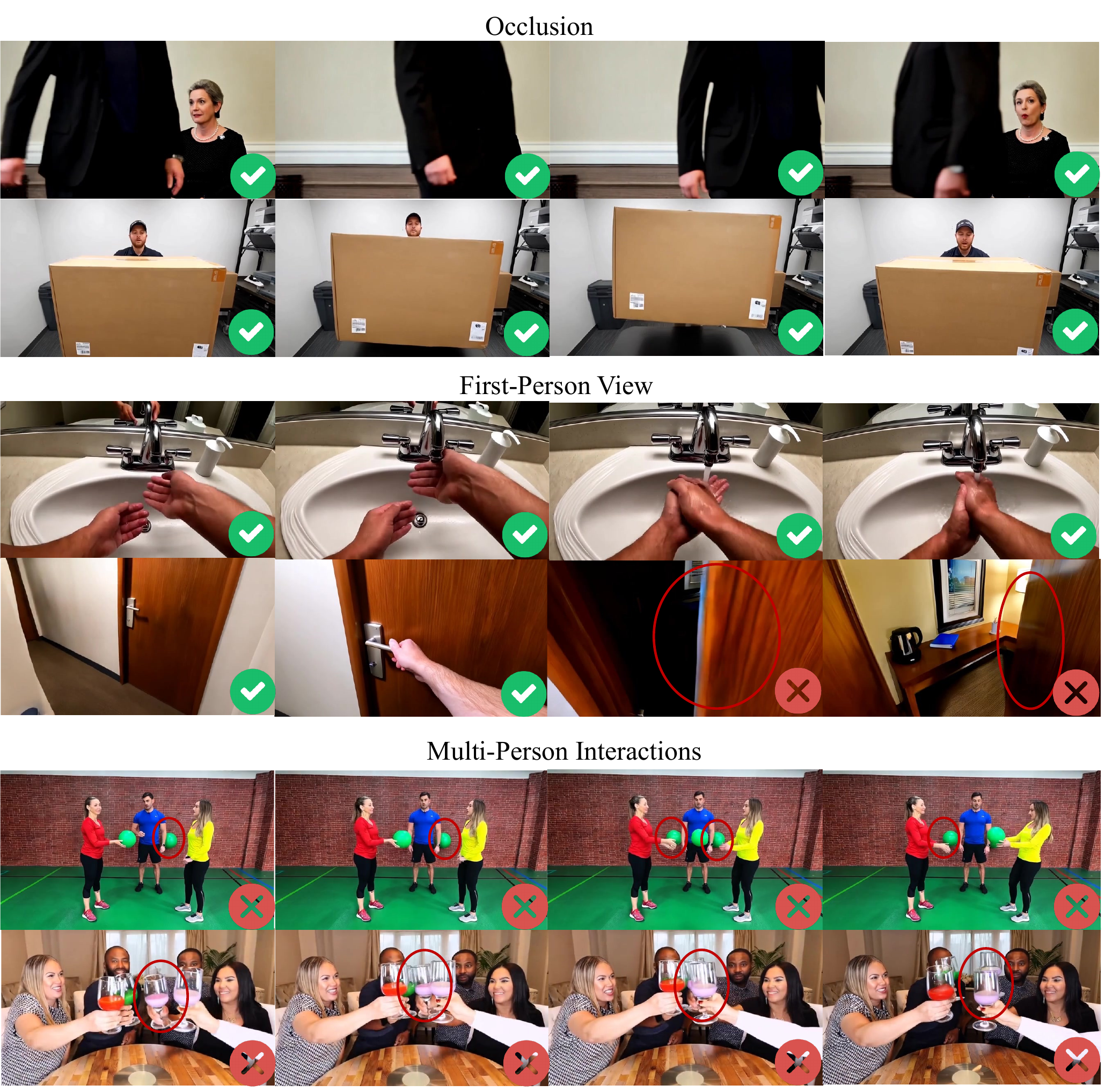}
    \caption{\revist{\textbf{Generalization under challenging scenarios.}
    We show representative results under occlusion, first-person views, and
    multi-person interactions. \nameofmethod{} remains robust to substantial
    occlusion and handles some first-person interactions, while more complex
    first-person actions and multi-person interactions may exhibit inconsistent
    object appearance, motion, or physical interactions. Green checks indicate
    successful cases, while red crosses and circles highlight representative
    failures and artifacts.}}
    \label{fig:limitation}
\end{figure}

\section{\revist{Limitations and Ethical Considerations}}
\label{sec:limitation}

Our work has two limitations.
\revist{First, generalization to less-covered and challenging scenarios remains
imperfect due to the limited diversity of large-scale paired audio-video data.
As illustrated in Fig.~\ref{fig:limitation}, \nameofmethod{} remains robust to
substantial occlusion and can handle some first-person interactions. However,
more complex first-person actions and multi-person interactions may still lead
to inconsistent object appearance, motion, or physically implausible
interactions.}
Second, although \nameofmethod{} enables few-step inference and reduces
streaming degradation, real-time high-resolution generation remains challenging.
Future work includes context compression and further step reduction.
\revist{As with other audio-video generative models, potential risks include
identity or voice misuse, privacy concerns, and biases inherited from the
training data. Responsible deployment should therefore consider identity/voice
consent and dataset provenance and licensing.}

\section{Conclusions}

We introduced \nameofmethod{}, a fast streaming framework for joint audio-video generation.
Our two-stage training first pretrains audio and video branches separately and then jointly trains them on paired audio-video data within a single, shared architecture, without adding separate cross-modal adapter modules.
To mitigate streaming degradation caused by train–inference context mismatch, we propose a unified dual-mode self-evolution scheme that shares weights between few-step and multi-step generation, improving quality while keeping inference efficient.
Experiments show that \nameofmethod{} matches or outperforms strong 50-step baselines using only 4--8 steps, enabling low-latency streaming generation.

\begin{acks}
This work was supported by NSFC
under Grants 62522607 and 62495061, and the Fundamental Research Funds for
the Central Universities (Nankai University).
\end{acks}

\bibliographystyle{ACM-Reference-Format}
\bibliography{egbib}

\end{document}